\documentclass{article}
\usepackage{spconf,amsmath,graphicx}
\usepackage[nolist]{acronym}
\usepackage{adjustbox}
\usepackage{amsmath}
\usepackage{multirow, array, booktabs,makecell,xspace,bm}
\usepackage{algorithm}
\usepackage{algorithmic}

\usepackage{amsmath}
\usepackage{makecell}
\usepackage{multirow}
\usepackage{multicol}
\usepackage{booktabs}
\usepackage{pifont}
\newcommand{\cmark}{\ding{51}}
\newcommand{\xmark}{\ding{55}}
\usepackage{arydshln}
\usepackage[safe]{tipa}
\usepackage{ifxetex}
\usepackage{algorithm}
\usepackage{algorithmic}
\usepackage{arydshln}
\usepackage{amssymb}
\usepackage{amsmath}

\usepackage[
backend=biber,
style=ieee,
citestyle=numeric-comp,
maxbibnames=3,
maxcitenames=3,
doi=false,isbn=false,url=false,eprint=false
]{biblatex}
\defbibheading{bibliography}{\begin{center}\textbf{6. REFERENCES}\end{center}}

\title{Exploring Phonetic Context-Aware Lip-Sync \\ For Talking Face Generation}
\name{Se Jin Park, Minsu Kim, Jeongsoo Choi, Yong Man Ro\sthanks{Corresponding Author. This work was partially supported by the National Research Foundation of Korea (NRF) grant funded by the Korea government (MSIT) (No. NRF-2022R1A2C2005529) and BK21 FOUR (Connected AI Education \& Research Program for Industry and Society Innovation, KAIST EE, No. 4120200113769).}}
\address{Integrated Vision and Language Lab. School of EE, KAIST, South Korea}
\begin{document}
\ninept
\maketitle
\begin{abstract}
Talking face generation is the challenging task of synthesizing a natural and realistic face that requires accurate synchronization with a given audio. Due to co-articulation, where an isolated phone is influenced by the preceding or following phones, the articulation of a phone varies upon the phonetic context. Therefore, modeling lip motion with the phonetic context can generate more spatio-temporally aligned lip movement. In this respect, we investigate the phonetic context in generating lip motion for talking face generation. We propose Context-Aware Lip-Sync framework (CALS), which explicitly leverages phonetic context to generate lip movement of the target face. CALS is comprised of an Audio-to-Lip module and a Lip-to-Face module. The former is pretrained based on masked learning to map each phone to a contextualized lip motion unit. The contextualized lip motion unit then guides the latter in synthesizing a target identity with context-aware lip motion. From extensive experiments, we verify that simply exploiting the phonetic context in the proposed CALS framework effectively enhances spatio-temporal alignment. We also demonstrate the extent to which the phonetic context assists in lip synchronization and find the effective window size for lip generation to be approximately 1.2 seconds.  
\end{abstract}

\section{Introduction}
Talking face generation aims to synthesize a photo-realistic portrait with lip motions in-sync with an input audio. Since the generation involves facial dynamics, especially of the mouth, viewers especially attend to the mouth region. Therefore, precisely aligning the mouth with the driving audio is critical for realistic talking face generation. In this paper, we focus on establishing audio-visual correlation for lip-syncing on arbitrary faces in the wild.

When a person utters the word `that', the corresponding lip movements vary based on the precedent and subsequent words. For instance, when saying `that boy', the lip configuration after pronouncing `that' tends to be closed, whereas it adopts a widely opened shape when saying `that girl'. This is because of the \textit{coarticulation}. \textit{Coarticulation} is a change in articulation of the current speech segment due to the neighboring speech. It arises due to the visual articulatory movements including the lips, tongue, and teeth being affected by the neighboring phones: a phone is either interchanged by a different phone or naturally blended in. With the realization of \textit{coarticulation}, phones have long been considered at the level of context such as triphones and quinphones in speech processing \cite{akita2005generalized, schwartz1985context, schwarz2006phoneme, golumbic2012temporal} as well as in speech-lip synchronization \cite{chung2020perfect,chen2021audio,kadandale2022vocalist,chung2017out}. 
% Therefore, for synthesizing authentic facial animation from speech, the phonetic context should be well addressed. 

% Coarticulation affected by preceding phones is referred to as preservatory while that affected by following phones as anticipatory.

Nevertheless, the phonetic context has not been sufficiently explored in talking face generation. The mainstream approach to processing the input speech has been formulating it as a short window of around 0.2 seconds to align the lip movement at the phone-level \cite{chen2018lip, song2018talking, jamaludin2019you, kr2019towards, vougioukas2020realistic, zhou2019talking, prajwal2020lip, park2022synctalkface}. However, since articulation is continuous and changes smoothly in context, independently building phone-level correlation without considering the context information results in ambiguities in the lip motion. Recent works \cite{zhang2021flow,zhang2021facial,yao2022dfa} employ transformer \cite{vaswani2017attention} to consider long-term context but they mainly aim to learn implicit attributes such as head pose and eye blinks from the context, and do not fully make use of the contextual information specifically for lip synchronization. 
In this paper, we aim to further exploit the phonetic context for lip sync generation and investigate the extent to which it contributes to realistic talking face generation.
% Since articulation varies upon the phonetic context, considering lip motion with the phonetic context can potentially generate more spatio-temporally aligned and stable lip movement.  

To effectively integrate the phonetic context in lip motion generation, we present Context-Aware Lip-Sync framework (CALS). CALS generates a talking face video of a target identity with context-aware lip motion synchronous with input audio. It consists of Audio-to-Lip module and Lip-to-Face module. The Audio-to-Lip module learns to predict lip motion units of the masked regions in the input audio at the phone-level. Each phone is mapped to lip motion units utilizing short-term and long-term relations between the phones. Hence, it translates audio to visual lip intermediate representations, associating the phonetic context while building the audio-lip correlation.  The Lip-to-Face module then leverages the contextualized lip motion units and integrates them with the identity to synthesize the face with context-aware lip synchronization. 
% Moreover, we analyze the effect of context on the lip sync generation and validate the effective context window.  

Our contributions are three-fold; (1) We propose Context-Aware Lip-Sync framework (CALS) that effectively exploits phonetic context in modeling lip-syncing for talking face generation. It explicitly maps each phone to contextualized lip motion units through masked learning, from which entire face frames within the context can be synthesized. (2) We explore to what extent the phonetic context contributes to lip generation and validate the effective context window size. (3) Through extensive experiments on LRW, LRS2, and HDTF datasets, we achieve a clear improvement in spatio-temporal alignment compared to the state-of-the-art methods. 

\section{Method}
We propose Context-Aware Lip-Sync framework (CALS) that generates a talking face video of a target identity with context-aware lip motion synchronous with a driving audio. CALS consists of Audio-to-Lip module and Lip-to-Face module. The Audio-to-Lip module learns to map audio to context-aware lip motion, namely lip motion units. The Lip-to-Face module synthesizes a talking face given the lip motion units and identity features, synthesizing the mouth region of the target identity. 

\subsection{Context-Aware Lip-Sync (CALS)}
\subsubsection{Audio-to-Lip Module} Audio-to-Lip module $f_{\theta}$ translates a given input audio into lip representation by considering the phonetic context. Motivated by the great success of masked prediction \cite{devlin2018bert,salazar2019masked,ghazvininejad2019mask,bao2020unilmv2,gulati2020conformer,kim2021lip,hong2022visual} in context modeling, we bring its concept into our learning problem to capture the phonetic context from the input audio when synthesizing the lip movements. Specifically, the audio input is processed into a sequence of mel-spectrograms which we denote as audio units $\mathbf A_{T}=\{a_{t}\}_{t=0}^{T}$, where $a_{t}$ is a frame-level mel-spectrogram at $t$-th frame. 
Then, we corrupt the audio units as follows: 
\begin{equation}
\mathcal{\Tilde{\mathbf A}}_{T} = r(\mathbf A_{T}, M),
\end{equation}
in which $M \subset [0,T]$ denotes the set of indices to be masked for the audio units $\mathbf A_{T}$, and $r$ is a masking function that zero-out the audio unit $a_{t}$ where $t \in  M$. As shown in Fig.1, we mask out a continuous sequence of $t_m$ audio units with probability $p$, without overlap between each segment. 
Audio-to-Lip module $f_{\theta}$ translates the corrupted audio units $\Tilde{\mathbf A}_{T}$ into contextualized lip motion units $\Tilde{\mathbf e}_{T}=\{\Tilde{e}_{t}\}_{t=0}^{T}$, 
\begin{equation}
\Tilde{\mathbf e}_{T}= f_{\theta}(\Tilde{\mathbf A}_{T}),
\end{equation}
where the Audio-to-Lip module $f_{\theta}$ is designed with transformer \cite{vaswani2017attention} architectures so the short-term and long-term context can be modeled. Then, it is guided to predict the corresponding ground truth lip motion units $\mathbf Z^{v}_{T}=\{z^{v}_{t}\}_{t=0}^{T}$  of the masked timestep $t \in M$ with the following loss function: 
\begin{equation}
\mathcal{L}_{a2l}= \sum_{t \in M} (\mathbf z^{v}_{t} - \Tilde{\mathbf e}_{t})^{2}.
 \end{equation}
% Setting the appropriate target is the key for context learning. In speech processing, vector quantized speech representation \cite{ling2020decoar}, discrete speech units via contrastive learning \cite{baevski2020wav2vec} and iteratively refined cluster ensembles \cite{hsu2021hubert} were used as the targets.
As our objective is producing synchronized lip motion from input audio by fully utilizing the phonetic context, the prediction target for our masked prediction should be carefully addressed. To handle this, we leverage a large-scale pre-trained visual encoder \cite{nagrani2020disentangled,chung2016out, Chung18b}, whose feature is set to our target for the masked prediction. The visual encoder is pre-trained using contrastive objectives between video frames and the corresponding audio frames. Thus, by predicting its visual features, the Audio-to-Lip module can focus on modeling synchronized lip movements from the input audio while minimizing the influence of speaker characteristics in the speech. The visual encoder extracts lip motion units from lip frames $\mathbf V_{T}=\{v_{t}\}_{t=0}^{T}$ in a sliding window of 5 frames, where the timestep $t$ is positioned at the center of the window. Hence, the lip motion units are discretized to exactly match the audio units. In order to predict the lip motion from the masked audio inputs, the Audio-to-Lip module has to first consider relations between the phones to predict the masked phones, and then map the phones to the synchronized visual lip features. As a result, the Audio-to-Lip module translates audio to visual lip intermediate representations, and associates phonetic context while establishing the audio-lip correlation. 

% The target lip motion units extracted from a pretrained visual encoder $E_{v}$ with the objective of audio-visual synchronization \cite{chung2016out}. 

%#################################################
\begin{figure}[t!]
	\begin{minipage}[b]{\linewidth}
		\centering		\centerline{\includegraphics[width=9cm]{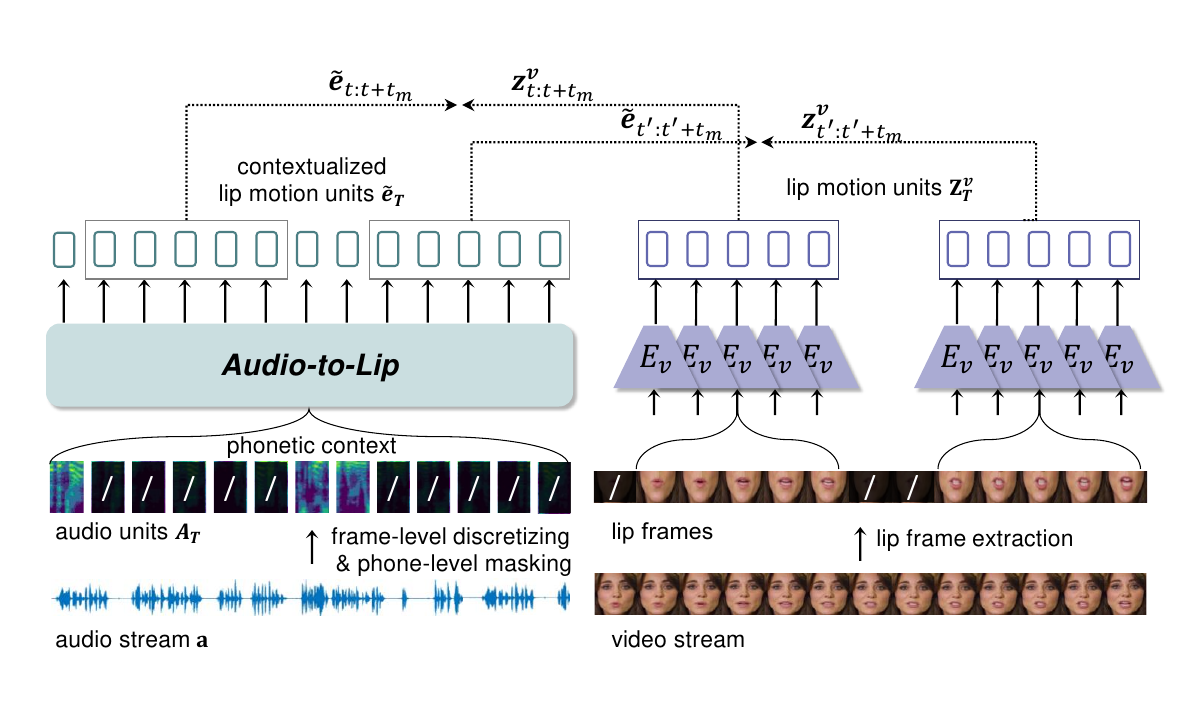}}
	\end{minipage}
	\vspace{-0.9cm}
	\caption{Audio-to-Lip module takes phone-level masked audio units as input and aims to predict the corresponding lip motion units of the masked regions.}
	\vspace{-0.4cm}
	\label{fig:1}
\end{figure}
%##################################################

\subsubsection{Lip-to-Face Module} 
Lip-to-Face module $f_{\phi}$ synthesizes face frames with the context-aware lip motion units drawn from the Audio-to-Lip module. Since it leverages the audio-lip correlation established in the Audio-to-Lip module, the Lip-to-Face module only has to focus on the synthesis part, and impose the dynamics of the lips onto the target identity. The Lip-to-Face module generates corresponding face frames $\mathbf{\hat{I}}_{T}$ within the context in parallel as follows: 
\begin{equation}
\mathbf{\hat{I}}_{T} = f_{\phi}({\mathbf e}_{T}, \mathbf{f}^{I}_{T}).
\end{equation}
$\mathbf{f}^{I}_{T}$ are corresponding identity features encoded by an identity encoder $E_{I}$. $E_{I}$ takes a random reference frame concatenated with a pose-prior along the channel dimension as in \cite{prajwal2020lip, park2022synctalkface}. The pose-prior is a target face with lower-half masked, which guides the model to focus on generating the lower-half mouth region that fits the upper-half pose. The lip motion units from the Audio-to-Lip module guide the Lip-to-Face module to generate lip shapes that are more distinctive to its phone in context. Moreover, as the lip motion units have attended to every other phones in the context, the dynamics are more temporally stable and consistent.

% \subsection{Video Synthesis}
% The Audio-to-Lip and Lip-to-Face modules are cascaded and trained end-to-end with the following generation loss. 
% \textbf{Reconstruction Loss.} 
% We apply L1 distance loss $\mathcal{L}_{recon}$ between the generated frames and ground truth frames.

% \textbf{Generative Adversarial Loss.} 
% We employ a GAN loss $\mathcal{L}_{gan}$ where a discriminator $D$ \cite{goodfellow2014generative} discriminates between real and fake images, penalizing unrealistic face generation.

\subsection{Discriminative Sync Loss}
In addition to the audio-lip correlation established in the Audio-to-Lip module, we further impose the alignment after the synthesis using a sync loss $\mathcal{L}_{sync}= d_{av} + d_{vv}$
% \begin{equation}
% \mathcal{L}_{sync} = d_{av} + d_{vv},
% \end{equation} 
which consists of an audio-visual sync loss $d_{av}$ and visual discriminative sync loss $d_{vv}$. 
Both of them use a discriminative sync module $(\mathcal{F}_a$, $\mathcal{F}_v)$ as feature extractors of the audio and visual modalities, pretrained on the target dataset with a multi-way matching loss \cite{chung2020seeing}. 

The audio-visual sync loss $d_{av}$ \cite{prajwal2020lip}  explicitly aligns the generated frames with the corresponding audio. It is a binary cross entropy of cosine similarity distance $d_{cos}$ between the generated video segment $\mathbf{\hat{I}}_{i}$ and the corresponding audio segment $\mathbf{a}_{i}$ as follows: 
\begin{equation}
    d_{av} = -\frac{1}{S}\sum_{i}^{S} \mathrm{log}      (d_{cos}(\mathcal{F}_a(\mathbf{a}_{i}), \mathcal{F}_v(\mathbf{\hat{I}}_{i}))).
\end{equation} 

In addition, to enforce discriminative lip motion within the context, we introduce visual discriminative sync loss $d_{vv}$. It compares one segment of the generated video to multiple segments of the corresponding ground truth video, thus guiding the generator to produce more distinct lip shapes from similar lip movements: 
\begin{equation}
d_{vv}=-\frac{1}{S}\sum_{i}^{S} \mathrm{log}\frac {\exp\big(d_{cos}(\mathcal{F}_v(\mathbf{\hat{I}}_{i}), \mathcal{F}_v(\mathbf{{I}}_{i}) )\big)}{\sum_{j}^{S} \exp\big(d_{cos}(\mathcal{F}_v(\mathbf{\hat{I}}_{i}), \mathcal{F}_v(\mathbf{{I}}_{j}) )\big)}. 
\end{equation}
$d_{vv}$ enforces discriminativeness of the lips by comparing directly against multiple numbers of the ground truth lips, aligning at the finer level in the visual domain. Also, since it learns to discriminate the matching lip motion within the context from which the lip unit has been produced, it enables more discriminative lip motion.  \\
\textbf{Total Loss.}
The CALS is trained end-to-end with the following loss: 
\begin{align}
\mathcal{L}= \lambda_{1}\mathcal{L}_{recon} + \lambda_{2}\mathcal{L}_{gan} + \lambda_{3}\mathcal{L}_{sync},
\end{align}
where $\mathcal{L}_{recon}$ is the L1 loss between the generated and ground truth frames, $\mathcal{L}_{gan}$ is the adversarial loss, and 
$\lambda_{n}$ is the weighting hyper-parameter. 

\section{Experiment}
\subsection{Experimental Settings}
\quad\,\,\,{\bf Dataset.}
We use three large-scale audio-visual datasets, LRW \cite{chung2016lip}, LRS2 \cite{afouras2018deep} and HDTF \cite{zhang2021flow} to train and evaluate the proposed method. LRW is a word-level dataset with over 1000 utterances of 500 different words. LRS2 is a sentence-level dataset with over 140,000 utterances. HDTF is a high-resolution dataset with more than 300 different speakers and 15.8 hours of approximately 10,000 utterances. 

{\bf Metric.}
We evaluate visual quality based on PSNR and SSIM, and sync quality based on LMD, LSE-D and LSE-C. LSE-C and LSE-D are the confidence score and distance score between audio and video features from SyncNet \cite{prajwal2020lip}. 

% \subsubsection{Comparison Methods.}
% We compare with five state-of-the-art methods of talking face generations: ATVGnet \cite{chen2019hierarchical}, Audio2Head \cite{wang2021audio2head}, PC-AVS \cite{zhou2021pose}, Wav2Lip \cite{prajwal2020lip}, and SyncTalkFace \cite{park2022synctalkface}. We use open-source codes to train on the target dataset.
% \vspace{-0.5cm}

%#################################################
\begin{figure}[t!]
	\centering
	\centerline{\includegraphics[width=8.5cm]{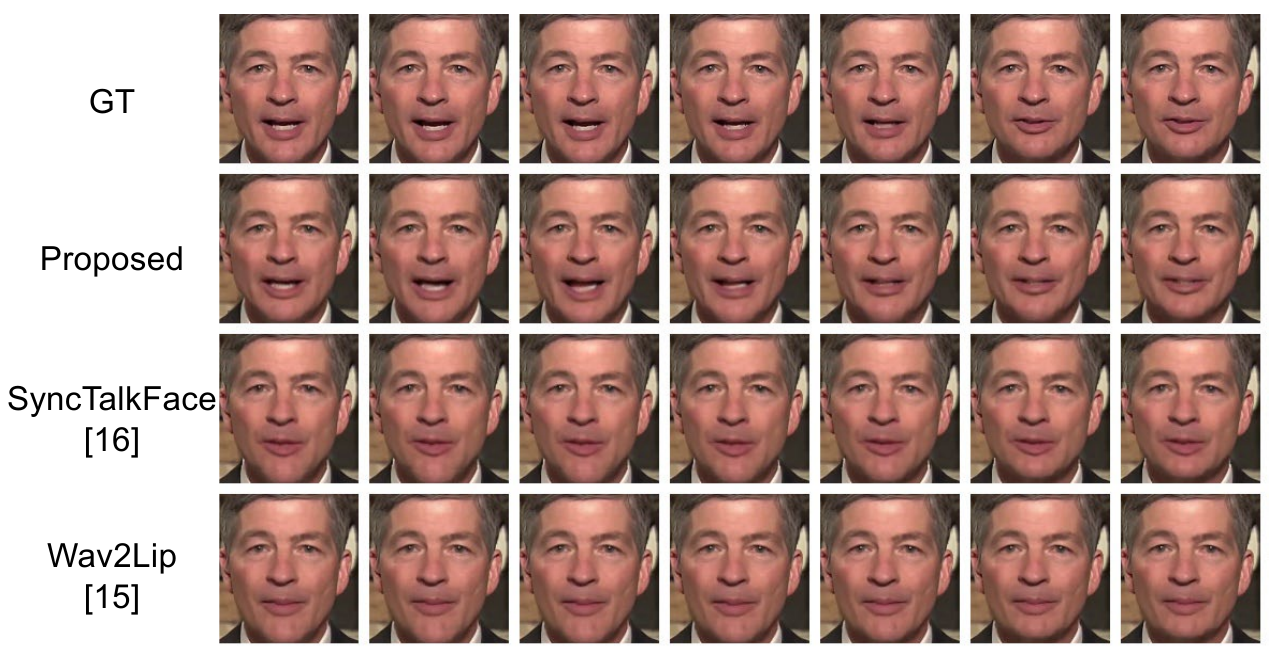}}
    \vspace{-0.5cm}
	\caption{Generation of frames with corresponding audio time-steps masked out. Please zoom in to see in detail.}
	\label{fig:5}
    \vspace{-0.1cm}
\end{figure}
%##################################################

%#################################################
\begin{figure}[t!]
	\centering
	\centerline{\includegraphics[width=8.7cm]{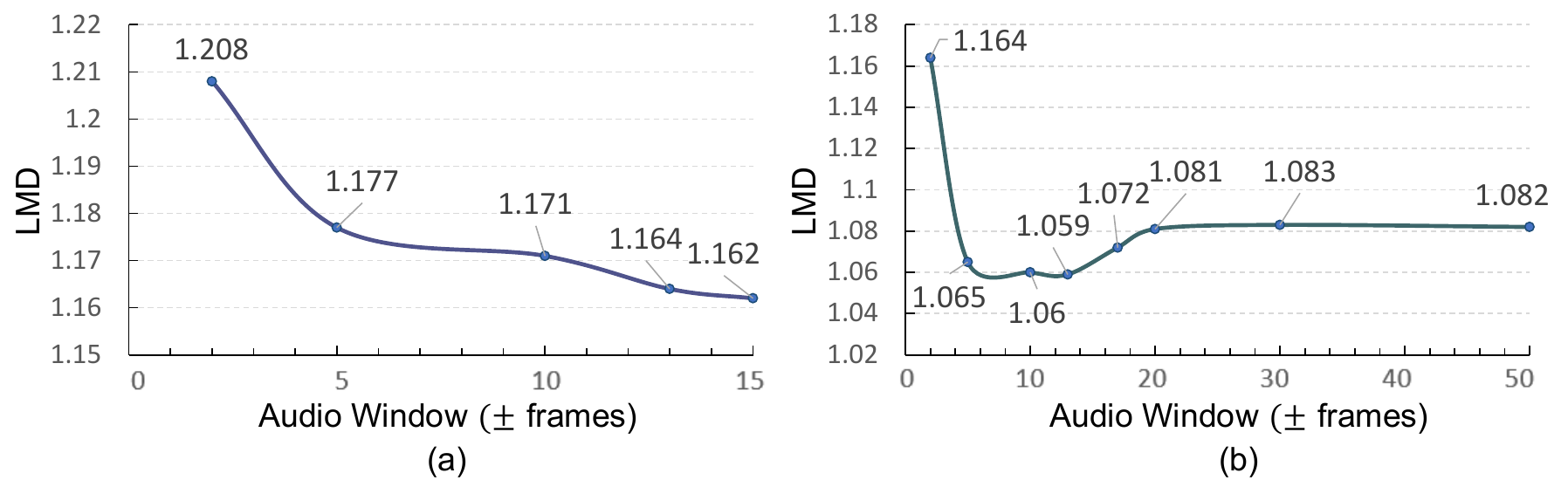}}
	\vspace{-0.3cm}
	\caption{LMD of the middle frame with varying audio window size on (a) LRW and (b) LRS2. }
	\label{fig:6}
\vspace{-0.2cm}
\end{figure}
%##################################################

{\bf Implementation Details.}
We process the videos into face-centered crops \cite{li2018dsfd} of size 128$\times$128 with 25fps. The audio is processed into audio units $a_{t}$ which are frame-level mel-spectrograms of size $80\times4$. We use sampling rate 16kHz, window size 400, and hop size 160. The Audio-to-Lip consists of 6 transformer encoder layers with hidden dimensions of 512, a feed-forward dimension of 2048, and 8 attention heads. We obtain the lip motion units from a pretrained sync module \cite{nagrani2020disentangled,chung2016out}. The sync module takes a sequence of 5 mouth frames and corresponding 0.2 seconds of audio centered at the $t$-th frame to extract a visual sync feature $z^{v}_{t} \in \mathbb{R}^{1024}$ and an audio sync feature $z^{a}_{t} \in \mathbb{R}^{1024}$. We mask out at the phone-level so $t_{m}$ is set to 5 and random 50$\%$ of the audio units are masked. The hyper-parameters are empirically set: $\lambda_{1}$ to 10, $\lambda_{2}$ to 0.07, and $\lambda_{3}$ to 0.03. We train on 8 RTX 3090 GPUs using Adam \cite{kingma2014adam} optimizer with a learning rate of $1\times10^{-4}$. 

\begin{table}[t]
	\renewcommand{\arraystretch}{1.2}
	\renewcommand{\tabcolsep}{1.2mm}
\centering
\vspace{-0.2cm}
\caption{Ablation study on the proposed method on LRS2.}
\resizebox{0.999\linewidth}{!}{
\begin{tabular}{ccccccccc}
\Xhline{3\arrayrulewidth}
\multicolumn{4}{c}{Proposed Method} & & & &\\ \cmidrule{1-4}
{Baseline} & {\makecell{A2L}} & \textbf{\makecell{$d_{av}$}} & \textbf{\makecell{$d_{vv}$}}  & {PSNR} & {SSIM} & {LMD} & {LSE-D} & {LSE-C}\\ \hline
\cmark & \xmark & \xmark & \xmark & 31.182 & 0.841 & 1.519 & 5.895 & 8.795\\
\cmark & \cmark & \xmark & \xmark & 32.419 & 0.870 & 1.311 & 5.623 & 9.144 \\
\cmark & \cmark & \cmark & \xmark & 32.501 & 0.867 & 1.064 & \textbf{5.204} & \textbf{9.421}\\
\cmark & \cmark & \cmark & \cmark & \textbf{32.603} & \textbf{0.876} & \textbf{1.056} & {5.337} & {9.225} \\
\Xhline{3\arrayrulewidth}
\end{tabular}}
\vspace{-0.2cm}
\label{table:4}
\end{table}

\begin{table}[]
    \renewcommand{\arraystretch}{1.2}
    \renewcommand{\tabcolsep}{2.1mm}
\centering
\caption{Human evaluation by MOS with 95\% confidence interval}
\vspace{0.1cm}
\resizebox{0.999\linewidth}{!}{ 
\begin{tabular}{cccc}
\Xhline{3\arrayrulewidth}
 \,\, Method \,\, & Visual Quality & Lip-Sync Quality & Realness \\
\hline
Ground Truth & {4.607 $\pm$ 0.080 }  & {4.687 $\pm$  0.071}  & {4.627 $\pm$ 0.081}\\
\cdashline{1-4}
Audio2Head \cite{wang2021audio2head} & {2.761 $\pm$  0.111}  & {2.721 $\pm$  0.134}  & {2.458 $\pm$  0.126}\\
PC-AVS \cite{zhou2021pose} & {2.567 $\pm$ 0.093}  & {3.109 $\pm$ 0.110}  & {2.458 $\pm$ 0.103}\\
Wav2Lip \cite{prajwal2020lip} & {2.975 $\pm$ 0.093}  & {3.557 $\pm$ 0.110}  & {3.109 $\pm$ 0.103}\\
SyncTalkFace \cite{park2022synctalkface} & {3.333 $\pm$ 0.102}  & {3.761 $\pm$ 0.100}  & {3.502 $\pm$ 0.102}\\
\hline
\textbf{Proposed} & {\bf{3.761 $\pm$ 0.086}} & {\bf {4.119 $\pm$ 0.088}}  & {\bf{3.940 $\pm$ 0.081}}\\
\Xhline{3\arrayrulewidth}
\end{tabular}}
\vspace{-0.2cm}
\label{table:2}
\end{table}

\subsection{Experimental Results}
\subsubsection{Context in Generation}
We perform two experiments to analyze to what extent context contributes to lip synchronization. First, we mask out 0.14s (7 frames) of the source audio and generate frames of the masked time steps. As shown in Fig.2, our work can still generate well-synchronized lips even with the absence of audio because it is able to attend to the surrounding phones and predict lip motion that fits into the masked region in context. In contrast, the previous works cannot generate correct lip movements because they do not consider surrounding phones. Such results verify that our model effectively incorporates context information in modeling lip movement of the talking face. 
In addition, we generate varying sizes of the audio window. We take a frame in the middle as a target frame and increase the size of the input audio window at the frame-level. The maximum audio window for the LRW is $\pm$15 and LRS2 is $\pm$78. The audio frames that lie outside the window are zero-padded and we measure the lip-sync quality of the generated target frame using LMD. As shown in Fig.3, on the LRW dataset, taking the entire audio window yields the best lip-sync performance. It reaches 1.162 in LMD at $\pm$15 frames. But when we further experiment on the LRS2 using wider audio windows, we find that the effect of temporal audio information reaches the optimum 1.059 at around $\pm$13 frames. It demonstrates that the audio context of around 1.2 seconds assists in resolving ambiguities in the lip shapes of phones, improving spatio-temporal alignment. 

\begin{table*}[t]
\renewcommand{\arraystretch}{1.2}
\renewcommand{\tabcolsep}{1.3mm}
\centering
\caption{Quantitative comparison with state-of-the-art methods on LRW, LRS2 and HDTF.}
\vspace{0.1cm}
\resizebox{1\linewidth}{!}{
\begin{tabular}{cccccccccccccccc} 
\Xhline{3\arrayrulewidth}
  & \multicolumn{5}{c}{LRW} & \multicolumn{5}{c}{LRS2} & \multicolumn{5}{c}{HDTF}  \\ \cmidrule(lr){2-6} \cmidrule(lr){7-11} \cmidrule(lr){12-16} 
 \,\, Method \,\, & PSNR & SSIM & LMD & LSE-D & LSE-C \,\, & \,\, PSNR & SSIM & LMD & LSE-D & LSE-C \,\, & \,\,  PSNR & SSIM & LMD & LSE-D & LSE-C \\
\hline
Ground Truth & {N/A} & {1.000}& {0.000} & {6.968} & {6.876}  \,\, & \,\,  {N/A} & {1.000} & {0.000} & {6.259} & {8.247} \,\, & \,\,  {N/A} & {1.000} & {0.000} & {7.508} & {7.128} \\
\cdashline{1-16}
Audio2Head \cite{wang2021audio2head} & {28.578} & {0.385} & {2.654} & {8.935} & {3.487} \,\, & \,\, {28.726} & {0.395} & {2.088} & {8.518} & {5.393} \,\, & \,\, {29.449} & {0.602} & {2.304} & {7.207}& {7.681} \\
PC-AVS \cite{zhou2021pose}& {30.257} & {0.746} & {1.989} & {6.502} & {7.438}  \,\, & \,\, {29.736} & {0.688} & {1.590} & {6.560} & {7.770} \,\, & \,\, {29.864} & {0.709} & {1.950} & {7.758} & {6.588}\\
Wav2Lip \cite{prajwal2020lip}& {31.831} & {0.882} & {1.437} & {6.617} & {7.237} \,\, & \,\,  {31.182} & {0.841} & {1.519} & {5.895} & {8.795} \,\, & \,\,  {32.354} & {0.873} & {1.595} & {7.272} & {7.343} \\
SyncTalkFace \cite{park2022synctalkface}& {32.887} & {0.894} & {1.322} & {7.023} & {6.591}  \,\, & \,\, {32.327} & {\bf0.881} & {1.069} & {6.350} & {7.929} \,\, & \,\, {32.682} & {0.883} & {1.381} & {7.931} & {6.406} \\
\hline
\textbf{Proposed} & {\bf33.219} & {\bf0.900} & {\bf1.183} & {\bf6.432} & {\bf7.463} \,\, & \,\, {\bf32.603} & {0.876} & {\bf1.056} & {\bf5.337} & {\bf9.225} \,\, & \,\, {\bf32.992} & {\bf0.895} & {\bf1.373} & {\bf6.850} & {\bf8.185} \\
\Xhline{3\arrayrulewidth}
\end{tabular}}
\label{table:1}
\end{table*}

%#################################################
\begin{figure*}[]
	\begin{minipage}[b]{1.0\linewidth}
		\centering
		\centerline{\includegraphics[width=18cm]{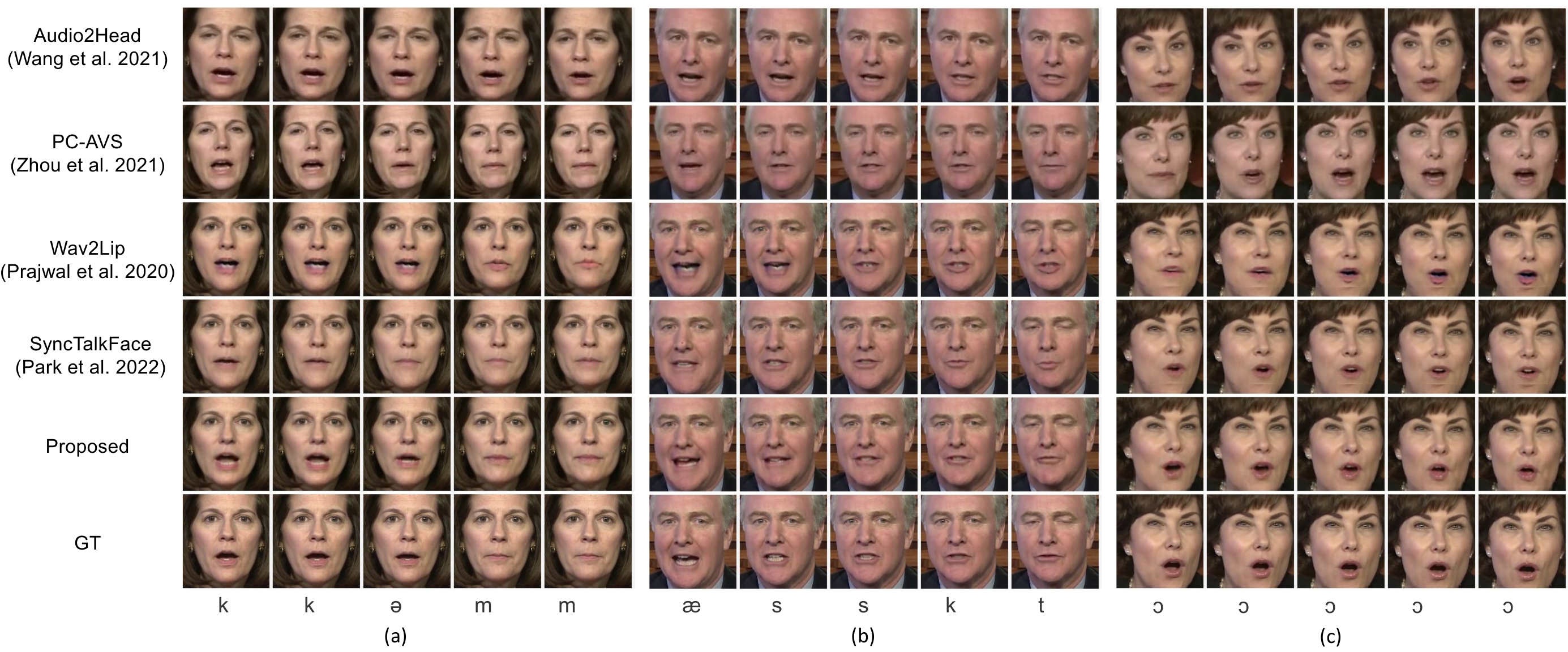}}
	\end{minipage}
    \vspace{-0.8cm}
	\caption{Qualitative comparison with state-of-the-art methods on HDTF (a) pronounces `com' in `combating', (b) `asked', (c) `a' in `all'. Phonemes corresponding to each frame are written under.}
    \vspace{-0.4cm}
	\label{fig:4}
\end{figure*}
%##################################################

\subsubsection{Analysis of the CALS}
We conduct an ablation to analyze the effect of each component of CALS in Table \ref{table:4}. We have set the baseline as Wav2Lip and implemented Audio-to-Lip module (A2L), $d_{av}$ and $d_{vv}$ one by one. When the phone-level audio encoder of the Wav2Lip is replaced by the A2L, the visual quality and sync quality are enhanced, where the SSIM improves by 0.029 and LSE-C by 0.349. 
Adding $d_{av}$ improves LSE-D and LSE-C because it acts on the audio-visual alignment which the two metrics measure. $d_{vv}$ further enhances PSNR, SSIM and LMD because it aligns in the visual domain. The three components altogether yield the highest performance overall. The result indicates that exploiting the phonetic context in the A2L scheme alone has the largest effect in improving the generation overall. 

% Replacing the phone-level audio encoder of the Wav2Lip with A2L greatly increases both visual and sync quality, demonstrating the efficacy of incorporating the phonetic context as well as exploiting lip motion as the intermediate representations. Adding $d_{av}$ increases LSE-D and LSE-C because it acts on the audio-visual alignment which the two metrics measure. $d_{vv}$ further enhances PSNR, SSIM and LMD because it aligns in the visual domain. The three components altogether yield the highest performance overall. 

\subsubsection{Quantitative Results}
We quantitatively compare the generation results with 4 state-of-the-art methods: Audio2Head \cite{wang2021audio2head}, PC-AVS \cite{zhou2021pose}, Wav2Lip \cite{prajwal2020lip}, and SyncTalkFace \cite{park2022synctalkface} on LRW, LRS2, and HDTF datasets in Table \ref{table:1}. On LRW and HDTF, our method achieves the best on all the metrics, especially outperforming on the lip-sync. Compared to other methods that involve feature disentanglement (PC-AVS), assistant module (SyncTalkFace, Wav2lip), and intermediate structural representations (Audio2Head), we validate that exploiting phonetic context in modeling lip motion is a more powerful scheme to achieve accurate lip synchronization.

% \begin{table}[t]
%     \renewcommand{\arraystretch}{1.2}
%     \renewcommand{\tabcolsep}{5mm}
% \centering
% \label{table:headings}
% \caption{Lip readability test of word prediction accuracy.}
% \resizebox{0.6\linewidth}{!}{ 
% \begin{tabular}{cc}
% \Xhline{3\arrayrulewidth}
%  \,\, Method \,\, & Accuracy (\%) \\
% \hline
% Ground Truth & {85.864}\\
% \cdashline{1-2}
% ATVGnet \cite{chen2019hierarchical} & 19.476$^\dagger$ \\
% Audio2Head \cite{wang2021audio2head} & {2.460} \\
% PC-AVS \cite{zhou2021pose} & {55.444}\\
% Wav2Lip \cite{prajwal2020lip} & {65.836}\\
% SyncTalkFace \cite{park2022synctalkface} & {70.036}\\
% \hline
% \textbf{Proposed} & {\bf 83.876}\\
% \Xhline{3\arrayrulewidth}
% \multicolumn{2}{l}{\small $^\dagger$836 failure cases as false predictions} 
% \end{tabular}}
% \label{table:3}
% \end{table} 

\subsubsection{Qualitative Results}
We qualitatively compare the generation results in Fig.4. Our method is most precisely aligned in spatio-temporal dimension and generates the most temporally stable lips, distinct to each phone in context. For example in (a), when pronouncing `k\textipa{@}m' in `combating', our method wide opens the mouth and gradually closes with smooth transitioning. On the other hand, other methods are temporally misaligned and discontinuous: Audio2Head fails to completely close the mouth at `m', Wav2Lip closes its mouth but with slightly projected lips, and SyncTalkFace fails to clearly open the mouth at phone `k'. The man in (b) is pronouncing `æskt’ in `asked for’. Our method is the only work that successfully captures the slightly projected lips at `t’ transitioning into `f’. Such results demonstrate that our method fully makes use of the phonetic context for temporally aligned and consistent lip synchronization.
% The lip shape of `t’ in the last frame assimilates to the lip shape of the preceding phone `f’.

\subsubsection{User Study}
We conducted a user study to compare the generation quality in Table 2. We generated 21 videos using each of the methods and asked 15 participants to rate the videos in terms of visual quality, lip-sync quality, and realness in the range of 1 to 5. Our method has the highest visual quality, lip-sync quality, and realness score, especially outperforming in the lip-sync quality, which is also reflected in the quantitative and qualitative results in Table 3 and Figure 4. 

% \subsubsection{Lip readability Test.}
% We take a pretrained state-of-the-art lip reading model \cite{kim2021multi} and measure word prediction accuracy on the generated LRW test set. As shown in Table \ref{table:3}, our proposed method outperforms, giving 83.876\% accuracy which is very close to the prediction accuracy of the ground truth videos 85.864\%. This shows the superiority of the proposed approach in synthesizing genuine talking face videos that are comparable to the real ones.  

\section{Conclusion}
We present Context-Aware Lip-Sync framework (CALS) that explicitly integrates phonetic context to learn lip-syncing for talking face generation. Audio-to-Lip module maps each phone to contextualized lip motion units and Lip-to-Face module synthesizes the entire face frames within the context in parallel. From extensive experiments, we validated that exploiting the phonetic context in the proposed CALS is a simple yet effective scheme to enhance the generation performance, specifically the lip-sync. In addition, we analyzed the extent to which the phonetic context assists in lip synchronization, complementing the missing audio, and verified the effective window size to be approximately 1.2 seconds. 

% From extensive experiments on the three large-scale public datasets, we demonstrate that the critical role of phonetic context in authentic lip-syncing, and we emphasize that the context affects the modeling of lip motion to a great extent.
% \bibliographystyle{IEEEbib}
% \bibliography{strings,refs}
\printbibliography[heading=bibliography]
\end{document}